\documentclass[preprint,12pt]{elsarticle}

\usepackage{amssymb}

\usepackage{optidef}
\usepackage{dsfont}

\journal{Nuclear Physics B}

\begin{document}

\begin{frontmatter}



\title{BORA: Bayesian Optimization for Resource Allocation}


\author[inst1]{Antonio Candelieri}

\affiliation[inst1]{organization={Department of Economics, Management and Statistics, University of Milano-Bicocca},
            addressline={Building U7, via Bicocca degli arcimboldi 8}, 
            city={Milan},
            postcode={20126}, 
            country={Italy}}

\author[inst1,inst2]{Andrea Ponti}
\author[inst3]{Francesco Archetti}

\affiliation[inst2]{organization={OAKS s.r.l.},
            city={Milan},
            postcode={20125}, 
            country={Italy}}
            
\affiliation[inst2]{organization={Department of Computer Science, Systems and Commuinication, University of Milano-Bicocca},
            addressline={Building U14, viale Sarca, 336}, 
            city={Milan},
            postcode={20126}, 
            country={Italy}}            

\begin{abstract}
Optimal resource allocation is gaining a renewed interest due its relevance as a core problem in managing, over time, cloud and high-performance computing facilities. Semi-Bandit Feedback (SBF) is the reference method for efficiently solving this problem. In this paper we propose (\textit{i}) an extension of the optimal resource allocation to a more general class of problems, specifically with resources availability changing over time, and (\textit{ii}) Bayesian Optimization as a more efficient alternative to SBF. Three algorithms for Bayesian Optimization for Resource Allocation, namely BORA, are presented, working on allocation decisions represented as numerical vectors or distributions. The second option required to consider the Wasserstein distance as a more suitable metric to use into one of the BORA algorithms. Results on (\textit{i}) the original SBF case study proposed in the literature, and (\textit{ii}) a real-life application (i.e., the optimization of multi-channel marketing) empirically prove that BORA is a more efficient and effective learning-and-optimization framework than SBF.
\end{abstract}


\begin{highlights}
\item Using Bayesian Optimization as a more effective and efficient approach than semi-bandit feedback to solve the optimal resource allocation problem
\item Extending the original formulation of the optimal resource allocation problem to a more general setting, with the aim to address new real-life problems
\item Evaluating the benefits of three different implementations of the proposed approach on a test problem (from the semi-bandit feedback literature) and a real-life setting (i.e., optimal resource allocation in multi-channel marketing).
\end{highlights}


\end{frontmatter}


\section{Introduction}\label{sec1}

\subsection{Motivation}
The reference problem considered in this paper is the \textit{optimal sequential resource allocation}. Assume that, at each time-step $t$, a decision-maker has to distribute a given \textit{budget} $\bar{b}^{(t)}$ over $m$ different \textit{options} with the aim to maximize the \textit{cumulative reward} of her/his decisions in $T$ time-steps. Formally, she/he wants to solve the following optimization problem:
\begin{maxi}|s|
{\mathbf{x}^{(t)}\in \mathbb{R}_+^m}
{\mathbb{E}\left[\sum_{t=1}^Tf\left(\mathbf{x}^{(t)}\right)\right]}
{}{}
\addConstraint{\sum_{i=1}^m x_i^{(t)} \leq \bar{b}^{(t)}}
\label{eq::P1}
\end{maxi}

where $f\left(\mathbf{x}^{(t)}\right)$ is the \textit{immediate reward} associated to the decision $\mathbf{x}^{(t)}$ made at time $t$. In some cases the constraint in (1) is considered as an equality instead of an inequality (i.e., all the budget must be used), as done later in this paper.

Optimal (sequential) resource allocation is a well known problem in operation research, with traditional examples such as inventory management, portfolio allocation, etc. \citep{ziukov2015literature,zhang2021spare,xidonas2020robust,gunjan2022brief}. Some approaches are related to stochastic optimization (e.g., multi-period and multi-stage stochastic optimization \citep{carpentier2015stochastic,bakker2020structuring}), but the most relevant literature for this paper is from Multi Armed Bandit (MAB)~\citep{slivkins2019introduction,lattimore2020bandit}, specifically the \textit{semi-bandit feedback} (SBF)~\citep{neu2013efficient,wen2015efficient,wang2018thompson,jourdan2021efficient,balcan2020semi,chen2021combinatorial}.
SBF has recently gained a renewed interest according to the technological advances and widespread usage of cloud and/or high performance computing facilities~\citep{vinothina2012survey,sonkar2016review,yousafzai2017cloud,wang2018thompson,thananjeyan2021resource}, where it is critical to optimally allocate computational resources (i.e., the budget) to different \textit{jobs} (i.e., options, arms) with the aim to maximize the number of successfully completed jobs over time. Analogously to MAB, also in SBF the learning process requires that every decision $\mathbf{x}^{(t)}$ must be taken by balancing between \textit{exploitation} (i.e., making the best decision according to the current knowledge about the problem) and $exploration$ (i.e., making a decision with the aim to reduce uncertainty instead of settling for the currently known best one). 
While MAB consists in selecting, at each step $t$, just one option (aka \textit{arm}) among the $m$ possible, SBF allows to simultaneously choose multiple arms, by allocating a different amount $x_i^{(t)}\geq 0$ of the available budget on each of them. The reward will depend on the amount of budget allocated on the different arms, altogether. It is also important to remark the difference with Combinatorial MAB (CMAB), where more than an arm can be pulled at each time-step (like in SBF), but decision variables, $x_i^{(t)}$, are binary insetad of numeric \citep{wang2018thompson}.

In the original formulation of \textit{optimal resource allocation through SBF}, proposed in \citep{lattimore2014optimal} and successively improved in \citep{dagan2018better}, the \textit{options} are computer processes (aka \textit{jobs}), the budget is constant over time and rescaled to 1 so that the constraint becomes $\sum_{i=1}^m x_i^{(t)}\leq 1$ and, consequently, $\mathbf{x}^{(t)}$ represents the allocation, in percentage, of the resources to the $m$ different computer processes at time-step $t$. Accordingly, the immediate reward is the number of successfully completed processes, whose individual probabilities of completion are given by Bernoulli distributions with unknown parameters $x_i^{(t)}/\nu_i$, that is $f\left(\mathbf{x}^{(t)}\right)=\sum_{i=1}^m \mathcal{B}\left(x_i^{(t)}/\nu_i\right)$. Since values $\nu_1, ..., \nu_m$ are unknown, the immediate reward is black-box. It is also important to clarify that, at the end of every time-step, the resources are replenished and all jobs are reset regardless of whether or not they completed in the previous time-step (i.e., there is not any \textit{inter-temporality} between consecutive decisions).

Indeed, the aim is to learn, as fast as possible, the values $\nu_1,...,\nu_m$, in order to choose $\mathbf{x}^{(t)}$ maximizing $\mathbb{E}\left[\sum_{t=1}^T \sum_{i=1}^m \mathcal{B}\left( x_i^{(t)}/\nu_i\right)\right]$, with $\sum_{i=1}^m x_i^{(t)}\leq1$. \\

Another well-known \textit{learning-and-optimization} framework is Bayesian Optimization (BO), a sample efficient sequential model-based global optimization method for expensive, multi-extremal, and possibly noisy functions~\citep{frazier2018bayesian,archetti2019bayesian,candelieri2021gentle}.
As MAB, BO is sequential and deals with the exploitation-exploration dilemma; the main difference among these two methods -- at least in their original formulations -- is that MAB optimizes over a discrete \textit{search space} (i.e., every decision is the selection of an arm among $m$), while BO optimizes over a continuous search space (i.e., every decision is a point of a box-bounded continuous space). However, the learn-and-optimize underlying nature of the two methods is so close that BO can be considered \textit{``a MAB with infinite arms''}, so that convergence proofs and results from MAB are often used to also prove convergence of BO methods, such as in~\citep{srinivas2012information}. More specifically, consider the following global optimization problem: 
\begin{equation}
    \underset{\mathbf{x} \in \Omega \subset \mathbb{R}^m}{\max} f(\mathbf{x})
\end{equation}

and denote with $r^{(t)}=f(\mathbf{x}^*)-f\left(\mathbf{x}^{(t)}\right)$ the so called \textit{instantaneous regret}, that is the difference between the actual (unknown) optimum $f(\mathbf{x}^*)$ and the immediate reward observed for the $t$-th decision. Then, denote with $R^{(T)}=\sum_{t=1}^T r^{(t)}$ the \textit{cumulative regret};~\citep{srinivas2012information} proved that, under suitable conditions, the cumulative regret of BO is sub-linear, meaning that $\underset{T\rightarrow \infty}{\lim} \frac{R^{(T)}}{T}=0$. 

This means that, if we could remove the constraint in (1) and impose $\mathbf{x}^{(t)}\in \Omega,\; \forall\; t=\{1,...,T\}$, then solving (2) via BO would lead to a solution for the \textit{unconstrained} version of the problem (1).
Consequently, a simple idea would be to address the problem (1) via Constrained BO (CBO)~\citep{gelbart2014bayesian,gelbart2015constrained,letham2019constrained,antonio2021sequential}, with $\bar{b}^{(t)}$ potentially changing at each time-step $t$.
Although possible, we are going to demonstrate that the CBO approach is not so efficient, and therefore we propose a novel method performing BO over the $m-1$ dimensional simplex, $\Delta_{m-1}$, requiring to work over univariate discrete probability measures instead of points in an $m$-dimensional box-bounded search space. The basic idea is that we are interested in searching for the optimal distribution of the currently available budget, independently on its amount. 

\subsection{Contributions}
\begin{itemize}
    \item extending the initial formulation of the optimal resource allocation problem \cite{lattimore2014optimal,dagan2018better} to the case of (\textit{i}) budget changing over time, that is $\bar{b}^{(t)}$, and (\textit{ii}) and equality constraint, that is $\sum_{i=1:m}x_i^{(t)}=\bar{b}^{(t)}$. The first aspect is a requirement emerging from real-life applications where the budget $\bar{b}^{(t)}$ cannot be directly decided by the decision-maker, but it comes out from other financial/commercial/managing decision processes. The second aspect represents the assumption that the entire budget must be allocated.
    \item a BO approach for optimal sequential resource allocation -- namely, BORA: Bayesian Optimization for Resource Allocation -- as an effective alternative to SBF.
    \item identification of classes of problems which can be effectively and efficiently solved through the proposed BORA algorithm(s).
\end{itemize}

\subsection{Related works}
A survey on Optimal Resource Allocation is given in \citep{patriksson2008survey}. Although not so recent, it provides a relevant overview about application domains and solving strategies. As far as the SBF methodology is concerned, many papers can be found in the recent literature, such as \citep{neu2013efficient,wen2015efficient,wang2018thompson,jourdan2021efficient,balcan2020semi,chen2021combinatorial}. With respect to its specific application to the optimal resource allocation problem, the most relevant references are \citep{lattimore2014optimal,lattimore2015linear,dagan2018better,verma2019censored,brandt2022finding}. Finally, other related works regard the topic of BO, especially Constrained BO (CBO), such as \citep{gelbart2014bayesian,gelbart2015constrained,letham2019constrained,antonio2021sequential}.

\subsection{Organization of the paper}
The rest of the paper is organized as follows. Section \ref{sec2} introduce the methodological background about SBF and GP-based BO. Section \ref{sec3} is devoted to detail the three proposed BORA algorithms. Section \ref{sec4} presents the two case studies considered, and then summarizes the empirical results. Finally, Section \ref{sec5} reports the most relevant conclusions, limitations, and perspectives.

\section{Background}\label{sec2}
\subsection{Optimal resource allocation via SBF in brief}
The difficulty underlying the optimal resource allocation problem is well described in \citep{lattimore2014optimal} and it could be summarized as follows: over-assigning resources to a job $i$ means that it will surely complete, but without providing any information about its difficulty (i.e., value of $\nu_i$), while under-assigning resources will lead to not complete the job, but it will allow to learn more about its difficulty, also saving resource for exploring other arms.
The real-life application motivating the idea of addressing optimal resource allocation problem via SBF was the cache allocation problem. In this problem, jobs are computer processes, with a job successfully completed, in a given time-step $t$, if there were no cache misses. As remarked in \citep{lattimore2014optimal}, other applications, such as load balancing in networked environments, or any other computing application characterized by precious resources (e.g., bandwidth, radio spectrum, CPU, etc.) to allocate among processes, can be addressed as optimal resource allocation via SBF.
Along with the introduction of the problem, \citep{lattimore2014optimal}, and successively \citep{dagan2018better}, propose an \textit{optimistic} SBF algorithm to solve the optimal resource allocation problem \ref{eq::P1}, with constant budget over time, rescaled to 1.

Their algorithm was inspired from the optimal policy which can be applied in the case of known $\nu_1,...,\nu_m$. This policy consists into fully allocate resource to the job with the smallest $\nu_i$ and then allocate the remaining resources to the next easiest job. Starting from that, in the proposed SBF algorithm, at each time-step $t$ the unknown $\nu_i$ is replaced with a high-probability lower bound \underbar{$\nu$}$_i^{(t)} \leq \nu_i$. This corresponds to the \textit{optimistic strategy}, which assumes that each job is as easy as reasonably possible. Naturally, learning a reliable confidence interval about every $\nu_i$ is delicate.

Indeed, under the assumption that the immediate reward function is $\sum_{i=1}^m \mathcal{B}\left(x_i^{(t)}/\nu_i\right)$, then it is non-differentiable at $x_i^{(t)}=\nu_i$, and for $x_i^{(t)}\geq\nu_i$ the job will always complete, providing little information about the job's difficulty. This issue is addressed by using the lower estimate of every $\nu_i$.
Furthermore, since $x_i^{(t)}$ varies over time, the observations of the successfully completed jobs are not identically distributed. This means that a standard sample average estimator is \textit{weak} in the sense that its estimation accuracy can be dramatically improved. On the contrary, the SBF algorithm estimates the lower and upper confidence bounds of each $\nu_i$, denoted with \underbar{$\nu$}$_i^{(t)}$ and $\bar{\nu}_i^{(t)}$, which are, respectively, non-decreasing and non-increasing. This is sufficient to guarantee that the SBF algorithm is numerically stable. The SBF algorithm and its improved version are reported, in pseudo-code, in \citep{lattimore2014optimal} and \citep{dagan2018better}, respectively.

\subsection{Gaussian Process based Bayesian Optimization in brief}

A Gaussian Process (GP) can be though as a collection of random variables, any finite number of which have a joint Gaussian distribution. A GP is completely defined by its mean and covariance function, respectively denoted with $\mu(\mathbf{x})$ and $k(\mathbf{x},\mathbf{x'})$, and can be compactly denoted with $\mathcal{GP}\big(\mu(\mathbf{x}),k(\mathbf{x},\mathbf{x'})\big)$~\citep{williams2006gaussian,gramacy2020surrogates}. Conditioning these two functions on a set of observations allows to use the GP as a probabilistic regression model to make predictions at any location $\mathbf{x}$.

More precisely, consider to have collected $t$ observations, denoted with $\mathbf{y}^{1:t}=\big\{y^{(i)}\big\}_{i=1:t}$, by having performed as many queries of $f(\mathbf{x})$ at the locations $\mathbf{X}^{1:t}=\big\{\mathbf{x}^{(i)}\big\}_{i=1:t}$. We use the notation $y^{(i)}$ because in the most general case the observations could be \textit{noisy}, that is $y^{(i)}=f\big(\mathbf{x}^{(i)}\big)+\varepsilon^{(i)}$, where $\varepsilon^{(i)}$ is assumed to be a zero-mean Gaussian noise, $\varepsilon^{(i)}\sim\mathcal{N}\big(0,\sigma^2_\varepsilon\big), \forall\; i\in\{1,...,t\}$).

Then, the GP's predictive mean and variance, conditioned to the $t$ performed queries, are respectively computed as follows:
\begin{equation}\label{eq:mu}
	\mu(\mathbf{x})=\mathbf{k}\big(\mathbf{x},\mathbf{X}^{1:t}\big)\big[\mathbf{K} - \sigma^2_\varepsilon\mathbf{I}\big]^{-1}\mathbf{y}^{(1:t)}
\end{equation}
\begin{equation}\label{eq:sd}
	\sigma^2(\mathbf{x})=k\big(\mathbf{x},\mathbf{x}\big) - \mathbf{k}\big(\mathbf{x},\mathbf{X}^{1:t}\big)\big[\mathbf{K} - \sigma^2_\varepsilon\mathbf{I}\big]^{-1} \mathbf{k}\big(\mathbf{X}^{1:t},\mathbf{x}\big)
\end{equation}

where $\mathbf{K}$ is a $t \times t$ matrix whose entries are $k_{ij}=k\big(\mathbf{x}^{(i)},\mathbf{x}^{(j)}\big)$, $\mathbf{k}(\mathbf{x},\mathbf{X}^{1:t})$ is the $t$-dimensional row vector $\big( k(\mathbf{x},\mathbf{x}^{(1)}), ..., k(\mathbf{x},\mathbf{x}^{(t)})\big)$, and $^\top$ is the transpose symbol.
The GP's predictive uncertainty is the posterior standard deviation, that is $\sigma(\mathbf{x})=\sqrt{\sigma(\mathbf{x})^2}$.

Before conditioning a GP to a set of observations, two priors must be provided, relatively to the mean and the covariance functions. Usually, the first is set to zero: this is not a limitation because the predictive mean (\ref{eq:mu}) will be not confined to this value. However, it is possible to incorporate explicit basis functions for expressing prior information on the function to be approximated by the GP model~\citep{williams2006gaussian}.
On the contrary, the covariance function is chosen among a set of possible \textit{kernel functions}, offering different modelling options with respect to structural properties of the function to be approximated, especially \textit{smoothness}~\citep{williams2006gaussian,archetti2019bayesian,frazier2018bayesian,gramacy2020surrogates}. In this paper the Squred Exponential (SE) kernel is used:
\begin{equation}
    k(\mathbf{x},\mathbf{x}') = \sigma_f \cdot e^{ -\frac{1}{2} \frac{\|\mathbf{x}-\mathbf{x}'\|^2}{\ell^2}}
\end{equation}

where $\sigma_f \in\mathbb{R}$ and $\ell\in\mathbb{R}^m_+$ are kernel's hyperparameters controlling, respectively, the vertical variation and the smoothness. When $\ell_i=\bar\ell,\;\forall\;i \in \{1,...,m\}$ the kernel is said \textit{isotropic}, otherwise a different smoothness is used along each dimension $i$ depending on $\ell_i$.\\

Since a GP provides both predictions and the associated uncertainty, it is said to be a \textit{probabilistic} regression model. This is crucial for dealing with the exploration-exploitation dilemma in BO.
Indeed, based on the current GP regression model, an \textit{acquisition function} (aka \textit{utility function} or \textit{infill criterion}) is optimized at every BO iteration with the aim to identify the location to query by balancing between
“trusting in prediction” (exploitation) and “giving a chance to uncertainty” (exploration). More precisely, the following auxiliary (aka ancillary or internal) optimization problem is solved:
\begin{equation*}
    \mathbf{x}^{(t+1)} = \underset{\mathbf{x}\in\Omega\subset\mathbb{R}^m}{\arg \max} \quad \mathcal{U}\left(\mathbf{x}; \left(\mu^{(t)}(\mathbf{x}),\sigma^{(t)}(\mathbf{x})\right)\right)
\end{equation*}

with $\mathcal{U}\left(\mathbf{x}; \left(\mu^{(t)}(\mathbf{x}),\sigma^{(t)}(\mathbf{x})\right)\right)$ the acquisition function, and the argument $\left(\mu^{(t)}(\mathbf{x}),\sigma^{(t)}(\mathbf{x})\right)$ making the dependence on the current GP regression model explicit. Several acquisition functions have been proposed offering alternative mechanisms to balance between exploration and exploitation \citep{frazier2018bayesian,archetti2019bayesian,candelieri2021gentle}: the well-known GP Upper Confidence Bound (GP-UCB) \citep{srinivas2012information} is used in this paper (and detailed in the following).

\section{The BORA algorithm(s)}\label{sec3}
In this section we present three different algorithms for solving optimal sequential resource allocation via BO. The firt -- named BORA$_1$ -- is basically CBO but with the budget constraint changing at each CBO iteration. The others two -- namely, BORA$_2$ and BORA$_3$ -- deal with the aforementioned constraint by remapping queries into the probability simplex. The difference between these two approaches consists in the GP's kernel: a SE kernel for BORA$_2$ and a Wasserstein-SE kernel -- which should be more suitable to work with elements of the probability simplex -- for BORA$_3$.

\subsection{BORA$_1$: BO over the search space under budget constraint}
At a generic time-step $t$, the following information are available to the BORA$_1$ algorithm:
\begin{itemize}
    \item the set of queried locations, $\mathbf{X}_{1:t}=\{\mathbf{x}^{(i)}\}_{i=1:t}$;
    \item the associated observations, $\mathbf{y}_{1:t}=\{y^{(i)}\}_{i=1:t}$;
    \item the new available budget $\bar{b}^{(t+1)}$.
\end{itemize}

To choose the next promising $\mathbf{x}^{(t+1)}$ -- balancing exploration and exploration -- BORA$_1$ performs the following two steps:
\begin{enumerate}
    \item it trains a GP depending on $\mathbf{X}_{1:t}$ and $\mathbf{y}_{1:t}$, by tuning the SE kernel's hyperparameters, $\sigma_f\in\mathbb{R}_+$ and $\ell \in \mathbb{R}^m_+$, via MLE maximization. The resulting GP works on the $m$-dimensional search space.
    
    \item it maximizes the UCB acquisition function with respect to an equality constraint, that is:
    \begin{equation*}
        \mathbf{x}^{(t+1)} = \underset{\mathbf{x} \in \Omega \subset \mathbb{R}^m}{\arg \max} \quad \mu^{(t)}(\mathbf{x}) + \sqrt{\beta^{(t)}} \sigma^{(t)}(\mathbf{x})
    \end{equation*}
    \begin{equation*}
        s.t. \quad \quad \sum_{i=1:m} x_i = \bar{b}^{(t+1)}
    \end{equation*}
\end{enumerate}

where $\beta$ is a parameter regulating the balance between pure exploitation (i.e., $\beta=0$) and pure random search (i.e., $\beta\rightarrow \infty$). Although \citep{srinivas2012information} originally proposed a logarithmic scheduling -- along with a convergence proof, under a limited number of queries -- more recently \citep{berk2021randomised} reported better performances by randomly sampling $\beta$ from a given distribution, outperforming the original scheduling on a range of synthetic and real-world problems.

\begin{figure}[h]
    \centering
    \includegraphics[scale=0.4]{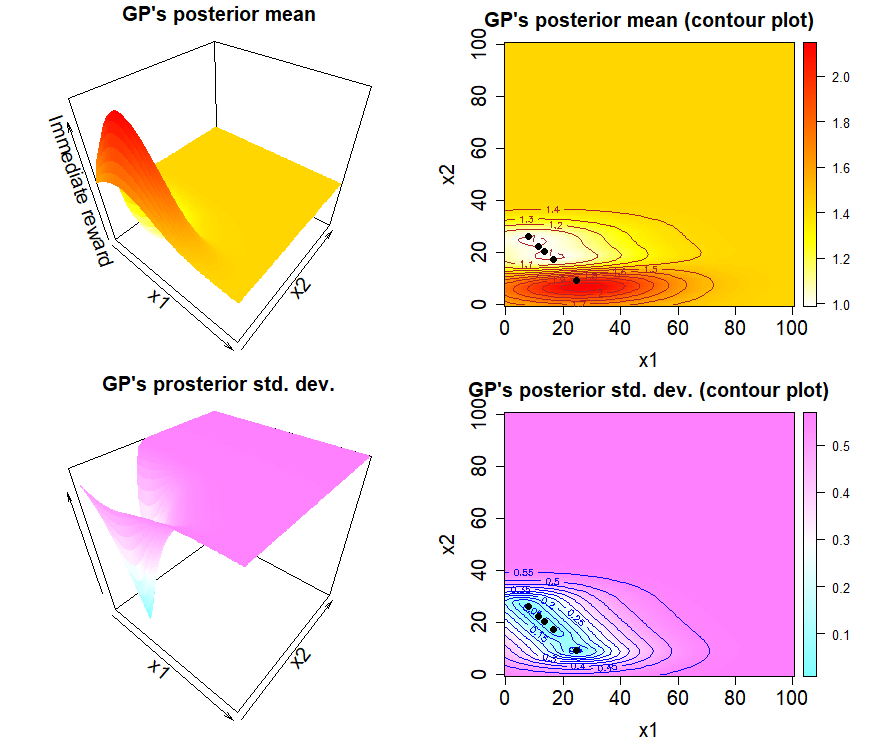}
    \caption{GP approximating the objective function, over the entire 2-dimensional search space $\Omega$ (i.e., two jobs), after 5 decisions and under constant budget over time. On the top, the GP's predictive mean, on the bottom, the GP's predictive standard deviation (aka predictive uncertainty).}
    \label{fig:2D_GP}
\end{figure}

To make BORA$_1$ clearer, we propose here a simple example, consisting of just $m=2$ arms, $\sum_{i=1}^2\mathcal{B}\left(x_i^{(t)}/\nu_i\right)$, $t=5$ already evaluated allocation decisions, and a constant budget equal to $33.9$ (i.e., randomly chosen). The BORA$_1$'s GP approximating the immediate reward function is represented in Figure \ref{fig:2D_GP}, and it is defined over the entire search space $\Omega=[0,100]^2$. In the upper part of Figure \ref{fig:2D_GP}, the GP's posterior mean, $\mu^{(t)}(\mathbf{x})$, is depicted, while in the lower part there is the GP's posterior standard deviation, $\sigma^{(t)}(\mathbf{x})$. All the already evaluated allocation decisions lay onto the line representing the constraint $x_1+x_2=33.9$.

\begin{figure}[h]
    \centering
    \includegraphics[scale=0.35]{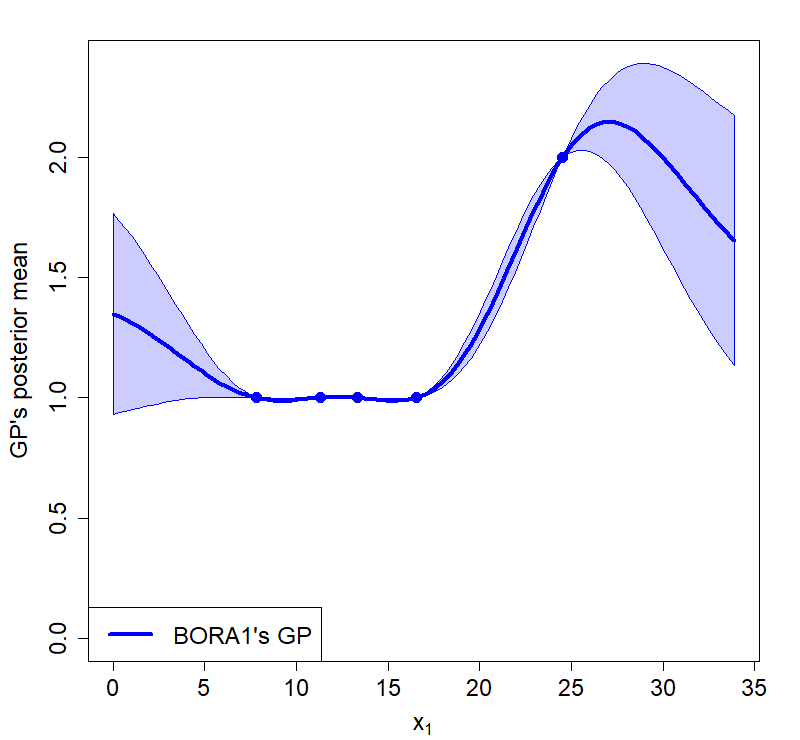}
    \caption{BORA$_1$'s GP (from Figure \ref{fig:2D_GP}), projected over the constraint $x_1+x_2=33.9$.}
    \label{fig:BORA1_GP}
\end{figure}

Although the GP is defined on the entire search space $\Omega$, the next promising allocation decision, $\mathbf{x}^{(t+1)}$, is selected according to the step 2 of the algorithm, meaning that it must be $x_1^{(t+1)}+x_2^{(t+1)}=33.9$. This means that we can restrict the UCB to that specific line, obtaining the 1-dimensional projection in Figure \ref{fig:BORA1_GP}. On the $x$-axis is the amount of budget allocated to the first arm, namely $x_1$, while $x_2$ is simply obtained as $33.9-x_1$.

Numeric values are taken from the first experiment analysed in this paper, related to the original SBF problem in \citep{lattimore2014optimal}, with $\nu_1=25$ and $\nu_2=50$. It is here important to remark that, after few iterations (i.e., 5 allocation decisions), BORA$_1$ has learned that allocating $x_1>=25$ increases the chance to maximize the immediate reward, exactly as the optimistic policy which SBF is based on.

\subsection{BORA$_2$: BO over the probability simplex via SE kernel}
The same set of information available to the previous algorithm is also available to BORA$_2$. As a relevant difference, BORA$_2$ introduces a further set, namely $\mathcal{A}_{1:t}$, which is basically a remapping of $\mathbf{X}_{1:t}$ into elements of the probability simplex
\begin{equation*}
    \Delta_{m-1}=\left\{\mathbf{a} \in \mathbb{R}^m_+, \sum_{j=1:m} a_j = 1 \right\}.
\end{equation*}

\noindent
More precisely, every element in $\mathcal{A}_{1:t}$ is computed as:
\begin{equation}
    \label{eq:map}
    \mathbf{a}^{(i)} = \mathbf{x}^{(i)}/\bar{b}^{(t)}.
\end{equation}

Thus, every $\mathbf{a}^{(i)}$ is the \textbf{weight vector} of an associated univariate discrete probability measure $\boldsymbol{\alpha}^{(i)}$ with fixed discrete support $\{1,..,m\}$.

Collapsing data points from $\Omega$ into univariate discrete probability distributions of the probability simplex $\Delta_{m-1}$, according to (\ref{eq:map}), leads to some important remarks:
\begin{itemize}
    \item the probability simplex $\Delta_{m-1}$ is a (linear) sub-space with dimensionality $m-1$ (i.e., one less than the original search space $\Omega$).
    \item two different $\mathbf{x}^{(i)}$ and $\mathbf{x}^{(j)}$ could collapse onto the same $\mathbf{a}$, depending on their associated $\bar{b}^{(i)}$ and $\bar{b}^{(j)}$ (e.g., $\mathbf{x}^{(i)}=(5,3,1)$, $\mathbf{x}^{(j)}=(10,6,2)$, $\bar{b}^{(i)}=9$, $\bar{b}^{(j)}=18$, leading to $\mathbf{a}^{(i)}=\mathbf{a}^{(j)}=(5/9,1/3,1/9)$). Therefore, observations $\mathbf{y}_{1:t}$ must be considered noisy, even if $f(\mathbf{x})$ would be noise-free over the original search space $\Omega$.
\end{itemize}

Finally, BORA$_2$ chooses the next promising location to query, $\mathbf{x}^{(t+1)}$, as follows:
\begin{enumerate}
    \item it trains a GP depending on $\mathcal{A}_{1:t}$ and $\mathbf{y}_{1:t}$, by tuning the SE kernel's hyperparameters, $\sigma_f\in\mathbb{R}_+$ and $\ell \in \mathbb{R}^m_+$, via MLE maximization. The resulting GP works on the simplex $\Delta_{m-1}$ (i.e., in equation (5), $\mathbf{x},\mathbf{x}'\in\Omega$ are replaced with $\mathbf{a},\mathbf{a}' \in \Delta_{m-1}$).
    
    \item it maximizes the UCB acquisition function over the simplex $\Delta_{m-1}$, that is:
    \begin{equation*}
        \mathbf{a}^{(t+1)} = \underset{\mathbf{a} \in \Delta_{m-1}}{\arg \max} \quad \mu^{(t)}(\mathbf{a}) + \sqrt{\beta^{(t)}} \sigma^{(t)}(\mathbf{a})
    \end{equation*}
    
    \item it maps $\mathbf{a}^{(t+1)}$ into \textbf{a uniquely associated} $\mathbf{x}^{(t+1)}$ according to the knowledge about $\bar{b}^{(t+1)}$, that is: $\mathbf{x}^{(t+1)}=\bar{b}^{(t+1)} \mathbf{a}^{(t+1)}$.
\end{enumerate}

\begin{figure}[h]
    \centering
    \includegraphics[scale=0.35]{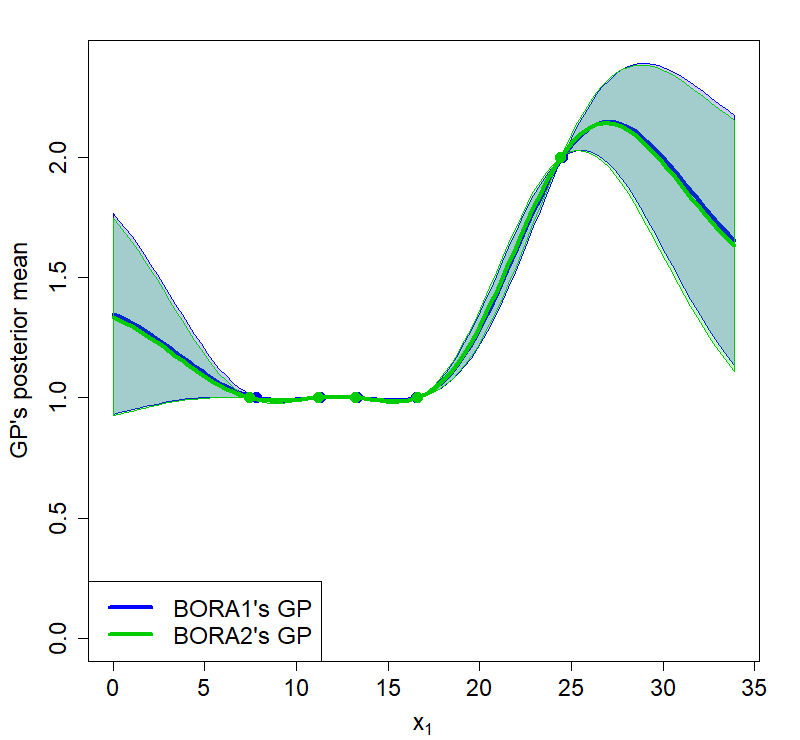}
    \caption{Comparing BORA$_1$'s GP (projected over the constraint $x_1+x_2=33.9$) and BORA$_2$'s GP. Since BORA$_2$'s GP works over the probability simplex $\Delta_{m-1}$, every distribution $\mathbf{a}^{(t)}$ was mapped to $\Omega$ according to $\mathbf{x}^{(t)}=\bar{b}^{(t)}\mathbf{a}^{(t)}$.}
    \label{fig:BORA2_GP}
\end{figure}

In Figure \ref{fig:BORA2_GP}, we compare the BORA$_1$'s and BORA$_2$'s GPs, given the same number of decisions already allocated (i.e., $t=5$). It is important to remark that, contrary to the BORA$_1$'s GP, that of the BORA$_2$ algorithm works on the probability simplex. Then, the probability distributions $\mathbf{a}^{(1)}, ..., \mathbf{a}^{(t)}$ are mapped into points $\mathbf{x}^{(1)}, ..., \mathbf{x}^{(t)}$ depending on the associated budgets $\bar{b}^{(1)}, ..., \bar{b}^{(t)}$, as previously mentioned. This mapping allows us to compare the two GPs in the picture. Although they result very similar in this case, this is not a general result (an example is reported in Appendix).

\newpage

\subsection{BORA$_3$: BO over the probability simplex via Wasserstein SE kernel}
The third algorithm, namely BORA$_3$ consists of the same steps described for the previous BORA$_2$. The relevant difference is in the GP's kernel: since we are working on the simplex $\Delta_{m-1}$, elements of this space are univariate discrete probability measures, so we decided to consider a more suitable distance to define an appropriate kernel between them. 
Recently, the interest on the Wasserstein distance \citep{villani2009wasserstein} and the associated Optimal Transport theory \citep{peyre2019computational} have been growing due to their successful application in many domains (e.g., imaging, signal processing and analysis, natural language process/generation, human learning, optimization, etc.)\citep{liu2021multi,wang2021imputation,zhang2021domain,zhang2021optimal,ponti2021new,ponti2021wasserstein,candelieri2022use}.
Very briefly, the Wasserstein distance allows to measure the difference between two probability distributions, $\boldsymbol{\alpha}$ and $\boldsymbol{\alpha'}$, independently on the values and the nature of their \textit{supports} (they can be both discrete, continuous, or one continuous and one discrete). The underlying idea consists in moving probability \textit{mass} with the aim to turn the first probability distribution, $\boldsymbol{\alpha}$, into the second, $\boldsymbol{\alpha}'$. Every movement has a cost which depends on a \textit{ground metric} defined over the supports. The Wasserstein distance is the minimum cost associated to the mass transportation turning the first probability measure into the second. The common notation is $\mathcal{W}_p(\boldsymbol{\alpha},\boldsymbol{\alpha}')$, with $0<p<\infty$ a parameter working on the ground metric.

We are specifically interested to the case of \textit{univariate discrete probability measures}. More precisely, a univariate discrete probability measure $\boldsymbol{\alpha}$ is defined by a unidimensional discrete support (i.e., $\{1,...,m\}$, in our study) and the associated \textit{weight vector} $\mathbf{a}\in \Delta_{m-1}:=\left\{\mathbf{a}' \in \mathbb{R}_+^m : \sum_{i=1}^m a_i'=1\right\}$. We remark that, in this paper, the weights vectors are obtained from the query points through the equation (\ref{eq:map}). 

With respect to our specific case study, we will use the important result in \citep{peyre2019computational}: although the Wasserstein distance is not generally Hilbertian -- that is, it cannot be efficiently approximated through a Hilbertian metric on a suitable feature representation of probability measures -- there exist special cases in which it can be computed in closed form and, in such cases, it is a Hilbertian measure.
In our study we consider: (\textit{i}) univariate probability measures with fixed support (all the distributions are defined over the support $\{1,...,m\}$) and  (\textit{ii}) a \textit{binary} ground metric, meaning that moving a unit of mass from $i$ to $j$ entails a cost of $1$ for every $j\neq i$, $0$ otherwise. Under this setting, $\mathcal{W}_p(\boldsymbol{\alpha},\boldsymbol{\alpha}')$ has a closed form, that is:
\begin{equation*}
    \mathcal{W}_p(\boldsymbol{\alpha},\boldsymbol{\alpha}') = \left[\frac{1}{2}\sum_{i=1}^m \vert a_i-a_i' \vert \right]^\frac{1}{p}
\end{equation*}

and it is therefore Hilbertian. This is important because, Hilbertian distances can be easily cast as radial basis function kernels:
for any Hilbertian distance $d$, it is indeed known that $e^{-d^q/\lambda}$ is a Positive Definite (PD) kernel, for any value $0 \leq q \leq 2$ and any positive scalar $\lambda$. Finally, we can define our Wasserstein-SE kernel:
\begin{equation}
    k_{\mathcal{W}SE}(\boldsymbol{\alpha},\boldsymbol{\alpha}')= \sigma_f \cdot e^{ -\frac{1}{2} \frac{ \mathbf{W}^2_p(\boldsymbol{\alpha},\boldsymbol{\alpha}')}{\lambda^2}}
\end{equation}

\begin{figure}[h]
    \centering
    \includegraphics[scale=0.35]{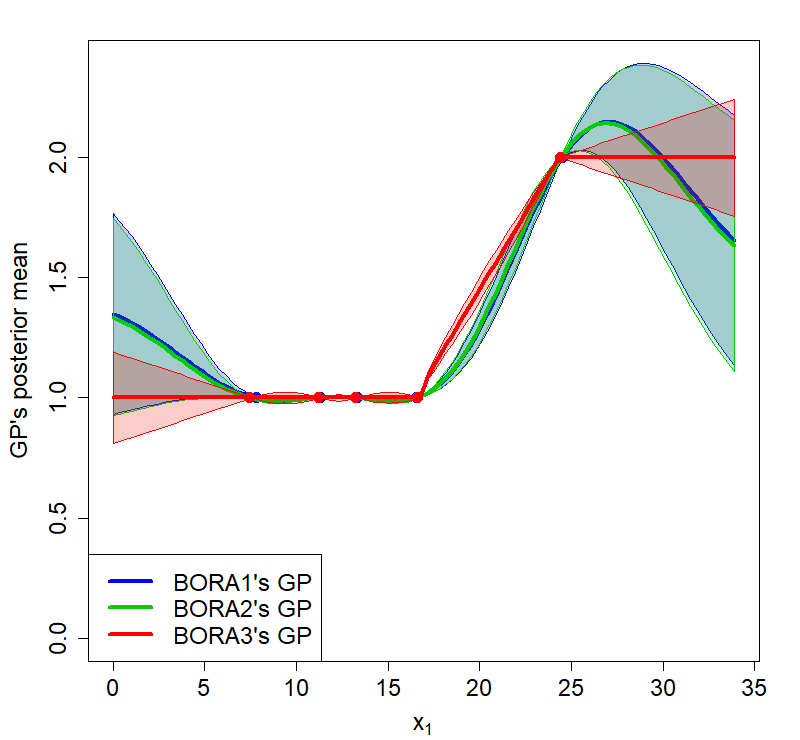}
    \caption{Comparing the GPs of the three BORA algorithms, after 5 allocation decisions and under a constant budget.}
    \label{fig:BORA3_GP}
\end{figure}

To conclude, the BORA$_3$ algorithm is the same of BORA$_2$, but it adopts $k_{\mathcal{W}SE}(\boldsymbol{\alpha},\boldsymbol{\alpha}')$ instead of $k_{SE}(\mathbf{x},\mathbf{x}')$. This modification is crucial because it affects the approximation of the immediate reward. Figure \ref{fig:BORA3_GP} compares the GPs of the three BORA algorithms, trained on the same set of $t=5$ allocation decisions. It is important to remind that the immediate reward, $\sum_{i=1}^2\mathcal{B}\left(x_i^{(t)}/\nu_i\right)$, is not-differentiable in $x_i^{(t)}=\nu_i$, and this property is well learned and modelled only by the BORA$_3$'s GP. Experiments will allow to empirically evaluate if this is sufficient to make BORA$_3$ the most performing algorithm.

\newpage

\section{Experiments and Results} \label{sec4}
\subsection{Case 1: optimal sequential allocation of computing resources}
This experimental case study is inspired by the one presented in \citep{lattimore2014optimal,lattimore2015linear,dagan2018better}, that consists in allocating an amount of computational resources, $x_i^{(t)}$, for $i=1,...,m$ jobs, at each step $t=1,...,T$. The aim is to maximize the number of successfully completed jobs over time, with the immediate reward function defined as follows:
\begin{equation}
    f\left(\mathbf{x}^{(t)}\right) = \sum_{i=1}^m \mathcal{B}\left(x_i^{(t)}/\nu_i\right)
\end{equation}

with parameters $\nu_i$ unknown to the learner/optimizer.
The relevant differences here are that (\textit{i}) the available budget can change over time, namely $\bar{b}^{(t)}$, and that (\textit{ii}) we consider an equality constraint instead of an inequality one. These are reasonable extensions emerging as requirements of real-life problems.

Thus, the problem considered in this use case can be formalized as follows:
\begin{maxi}|s|
{\mathbf{x}^{(t)}\in \mathbb{R}_+^m} {\mathbb{E}\left[\sum_{t=1}^T f\left(\mathbf{x}^{(t)}\right)\right]}
{}{}
\addConstraint{\sum_{i=1}^m x_i^{(t)} = \bar{b}^{(t)}, \quad \forall\; t}
\label{eq::P1}
\end{maxi}

As in \citep{lattimore2014optimal}, we first consider $m=2$. We have divided the analysis into two scenarios:
\begin{itemize}
    \item \textbf{Case 1.a}: the budget $\bar{b}^{(t)}$ is randomly chosen and kept constant over time, that is $\bar{b}^{(t)}=\bar{b},\;\forall\;t=1,...,T$. This means that all the values can be rescaled in order to have $\bar{b}=1$, therefore, the only difference with respect to the original SBF setting is the equality constraint.
    \item \textbf{Case 1.b}: the budget $\bar{b}^{(t)}$ changes over time with values uniformly sampled in $[10,100]$.
\end{itemize}

In both the sub-cases, $T=100$, $\nu_1=25$, and $\nu_2=50$. To mitigate the effect of randomness, 5 independent runs have been performed for each algorithm and each case. For every run, all the algorithms share the
same setting (i.e., immediate reward’s parameters, and budgets).\\

\begin{figure}[h]
    \centering
    \includegraphics[scale=0.35]{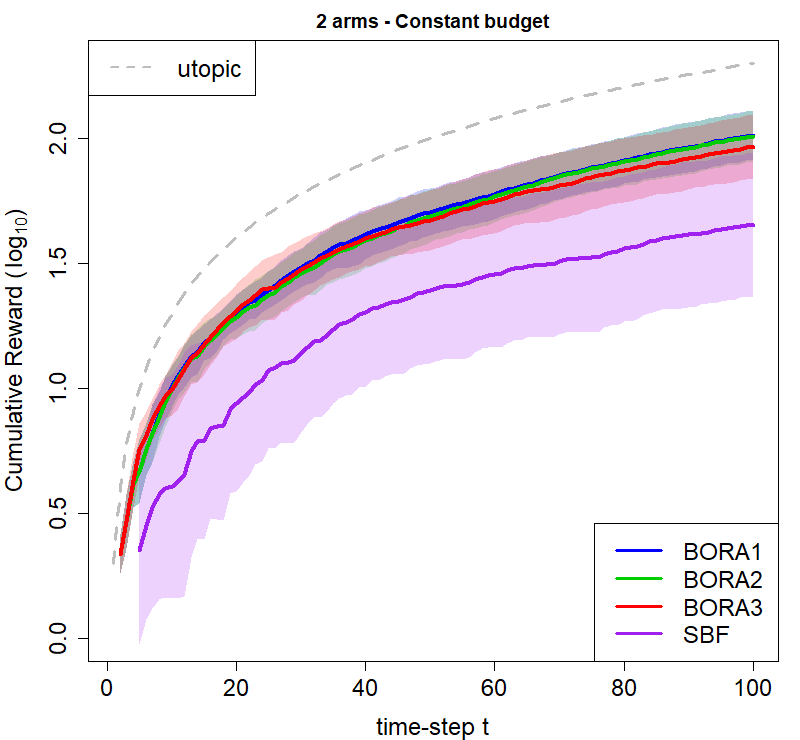}
    \caption{\textbf{Case 1.a (m=2)} results: cumulative reward (log$_{10}$) of each algorithm (mean and standard deviation over 5 independent runs). The dashed grey curve represents the \textit{utopic} performance (i.e., all the jobs always complete successfully).}
    \label{fig:case1a}
\end{figure}

Figure \ref{fig:case1a} shows the cumulative reward for the different algorithms with respect to the \textbf{Case 1.a}: the averages (solid lines) and standard deviations (shaded areas) computed over the 5 independent runs are depicted, separately for each algorithm.
It is evident that the three BORA algorithms offer similar performances, significantly outperforming SBF \citep{lattimore2014optimal,dagan2018better}. Values for SBF starts after few iterations because it initially collected zero as immediate reward (thus the log$_{10}$ operation led to ``\textit{not-a-number}'', even if the procedure suggested in \citep{lattimore2014optimal} to initialize the lower bounds \underbar{$\nu$}$_i$ was used).

\newpage

\begin{figure}[h]
    \centering
    \includegraphics[scale=0.35]{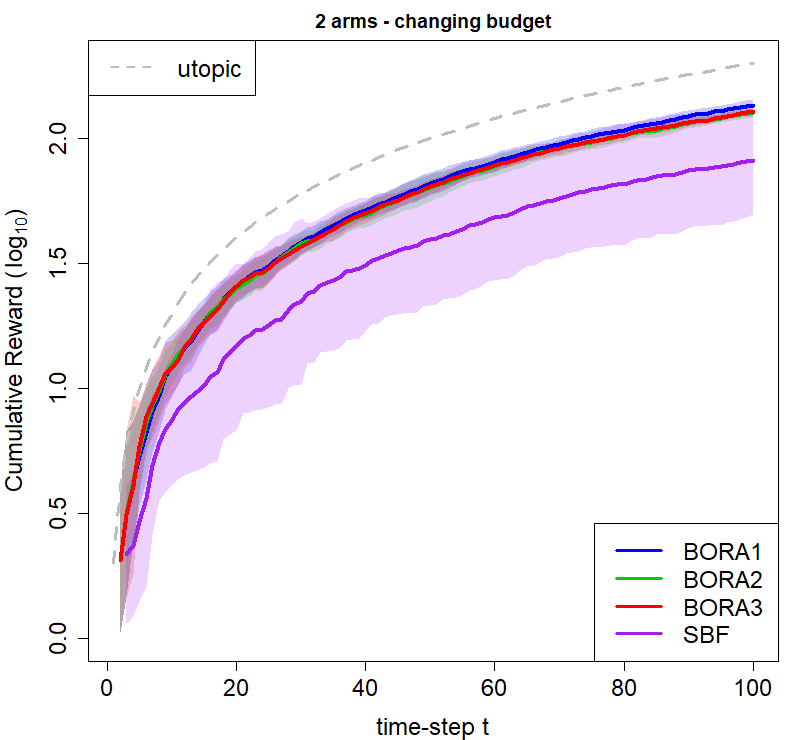}
    \caption{\textbf{Case 1.b (m=2)} results: cumulative reward (log$_{10}$) of each algorithm (mean and standard deviation over 5 independent runs). The dashed grey curve represents the \textit{utopic} performance (i.e., all the jobs always complete successfully).}
    \label{fig:case1b}
\end{figure}

Figure \ref{fig:case1b} shows the cumulative reward for the different algorithms with respect to the \textbf{Case 1.b}. Again, the three BORA algorithms show similar performances, while SBF was significantly under-performing.

However, it is important to remark that all the algorithms have shown an increase in the cumulative reward with respect to the previous Case 1.a (i.e., all the curves are closer to the utopic one, that is the curve of the cumulative reward with all the jobs always completed). This could be due to higher budgets, in some time-step, than the one (constant) in the Case 1.a, but it also empirically proves that all the algorithms are able to learn from experience and that \textit{diversity}, implied by the \textit{dynamic} setting (i.e., a changing budget over time), improves their \textit{generalization} capabilities.

\newpage

\begin{figure}[h]
    \centering
    \includegraphics[scale=0.35]{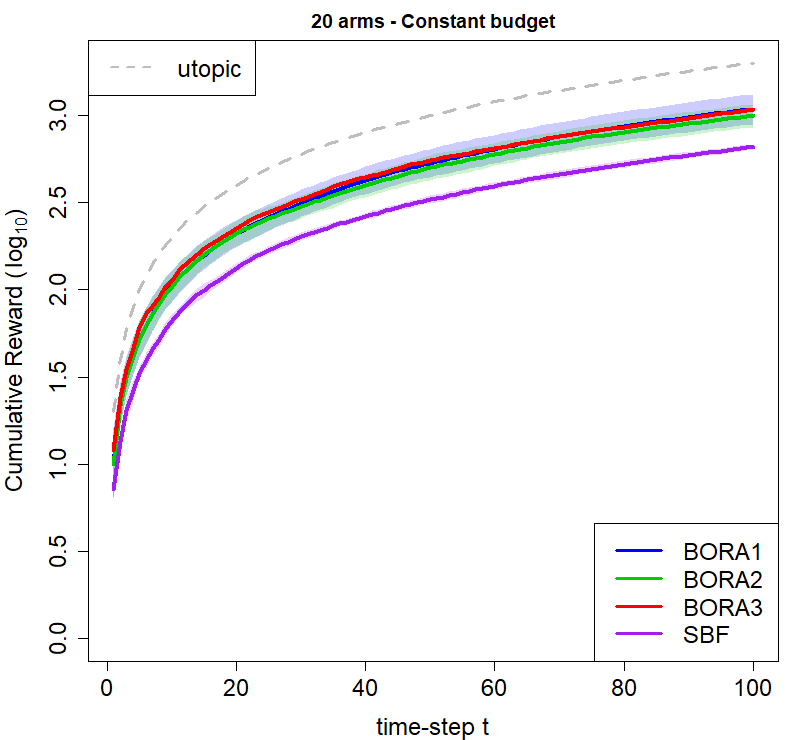}
    \caption{\textbf{Case 1.a (m=20)} results: cumulative reward (log$_{10}$) of each algorithm (mean and standard deviation over 5 independent runs). The dashed grey curve represents the \textit{utopic} performance (i.e., all the jobs always complete successfully).}
    \label{fig:case1a_20d}
\end{figure}

Then, we have decided to analyze which could be the impact of the dimensionality of the problem (i.e., the number $m$), on the algorithms' performances. Thus, the two cases have been addressed again by considering 20 arms, leading to a 20-dimensional search space. This choice is not casual: it is well known that BO scales poorly with the dimensionality of the problem, typically starting from 15-20 dimensions.
Figure \ref{fig:case1a_20d} shows the cumulative reward of the compared algorithms for the \textbf{Case 1.a with m=20}. Results confirm those of the previous Case 1.a (with $m=2$): the three BORA algorithms offer similar performances and outperform SBF. Interestingly, the standard deviation of the performances decreased for all the algorithms.

\newpage

\begin{figure}[h]
    \centering
    \includegraphics[scale=0.35]{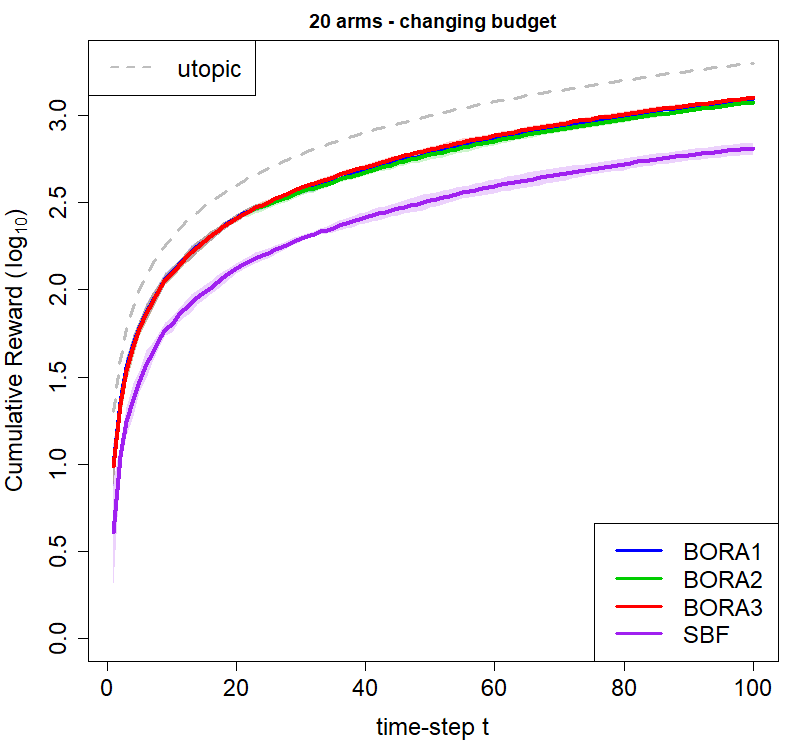}
    \caption{\textbf{Case 1.b (m=20)} results: cumulative reward (log$_{10}$) of each algorithm (mean and standard deviation over 5 independent runs). The dashed grey curve represents the \textit{utopic} performance (i.e., all the jobs always complete successfully).}
    \label{fig:case1b_20d}
\end{figure}

Figure \ref{fig:case1b_20d} shows the cumulative reward of the compared algorithms for the \textbf{Case 1.b with m=20}. Results confirm those of the previous Case 1.b (with $m=2$): the three BORA algorithms have similar performances and outperform SBF.

\newpage

\begin{figure}[h]
    \centering
    \includegraphics[scale=0.35]{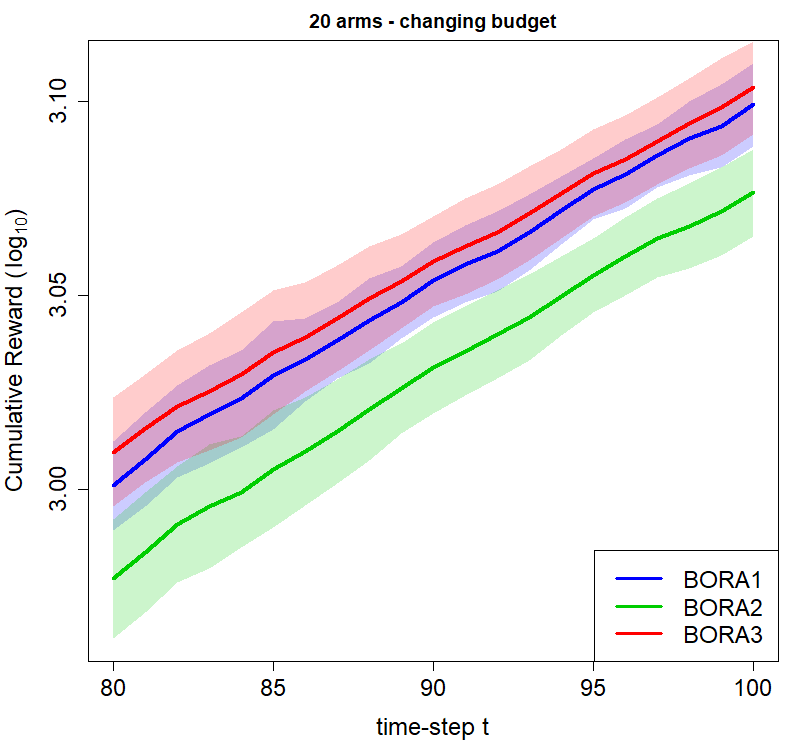}
    \caption{\textbf{Case 1.b (m=20arms)}, zoom-in on the last 20 iterations.}
    \label{fig:case1b_20d_zoom}
\end{figure}

Finally, in the \textbf{Case1.b with $m=20$}, a significant difference has been observed between the three BORA algorithms, as zoomed-in in Figure \ref{fig:case1b_20d_zoom}.
Indeed, BORA$_3$ was, on average, better than the other two BORA algorithms, and both BORA$_1$ and BORA$_3$ were significantly better than BORA$_2$. More precisely, the average cumulative reward, at the end of the 100 iterations, was $1256.50$, $1192.49$, and $1269.21$, for BORA$_1$, BORA$_2$, and BORA$_3$, respectively (all with standard deviation around $1.025$).

\newpage

\subsection{Case 2: optimal sequential allocation of marketing budget across multiple channels}
The second case study is inspired to the real-life problem of allocating marketing budget across multiple channels.
\textit{Multi-channel marketing} communicates a product or service’s value using the unique strengths of different marketing channels (e.g., email, websites, social media, display adverts, etc.). A strategy which ties together campaigns from multiple channels creates opportunities for more impacting messages that are mindful of the customer journey. However, more channels require a more efficient and effective management of economic resources to allocate on each channel to maximize revenues.\\

The problem is still formalized as follows:
\begin{maxi}|s|
{\mathbf{x}^{(t)}\in \mathbb{R}^m} {\mathbb{E}\left[\sum_{t=1}^T f\left(\mathbf{x}^{(t)}\right)\right]}
{}{}
\addConstraint{\sum_{i=1}^m x_i^{(t)} = \bar{b}^{(t)}, \; \forall\; t}
\label{eq::P1}
\end{maxi}

but now the immediate reward is $f\left(\mathbf{x}^{(t)}\right)=\sum_{i=1}^m\eta_i^{(t)} x_i^{(t)}$, with $\eta_i^{(t)}$ representing the unknown return of the investment $x_i^{(t)}$ on the channel $i$ at the step $t$. Exactly as for the previous case study, the immediate reward's parameters (i.e., $\eta_i^{(t)}$, in this case) are unknown to the learner, while at each step $t$ -- and before taking the decision $\mathbf{x}^{(t)}$ -- the available budget $\bar{b}^{(t)}$ is known.
For our experiments we have considered a time horizon $T=100$ and $m=15$ different channels. Budgets and returns are chosen as follows:
\begin{itemize}
    \item $\bar{b}^{(t)} \sim \mathcal{N}(50,10)$;
    \item $\eta_i^{(t)} \max\{0,\sim \mathcal{N}(\mu_{\eta_i},\sigma_{\eta_i})\}$, where $\mu_{\eta_i}$ and $\sigma_{\eta_i}$ are, themselves, sampled as $\mu_{\eta_i} \sim \mathcal{U}(0,1)$ and $\sigma_{\eta_i} \sim \mathcal{U}(0,0.2)$. All the specific numeric values can be checked at the shared code repository (as later detailed in the section titled ``Code availability'').
\end{itemize}

It is also important to remark some crucial differences with respect to the previous Case1:
\begin{itemize}
    \item the immediate reward -- still unknown to the learner -- is now a generic linear function (instead of a sum of zeros and ones);
    \item due to the nature of the immediate reward, the SBF algorithm cannot be used -- \textit{as-is} -- to solve this case study, because it relies on the assumption that the terms of the sum, in the immediate reward, are binary (i.e., $0$ or $1$);
    \item the three BORA algorithms can be applied to this new case study without any modification: they are able to model -- and optimize -- the immediate reward independently on its nature, and they require just the value of the reward instead of the knowledge of its individual terms.
\end{itemize}

Again, 5 independent runs have been performed for each BORA algorithm, to mitigate the effect of randomness. For every run, all the algorithms share the same setting (i.e., objective function's parameters and budgets).

As for the previous case study, two different sub-cases are considered, that are: \textit{constant} budget (i.e., \textbf{Case 2.a}) and \textit{changing} budget (i.e., \textbf{Case 2.b}).

\begin{figure}[h]
    \centering
    \includegraphics[scale=0.35]{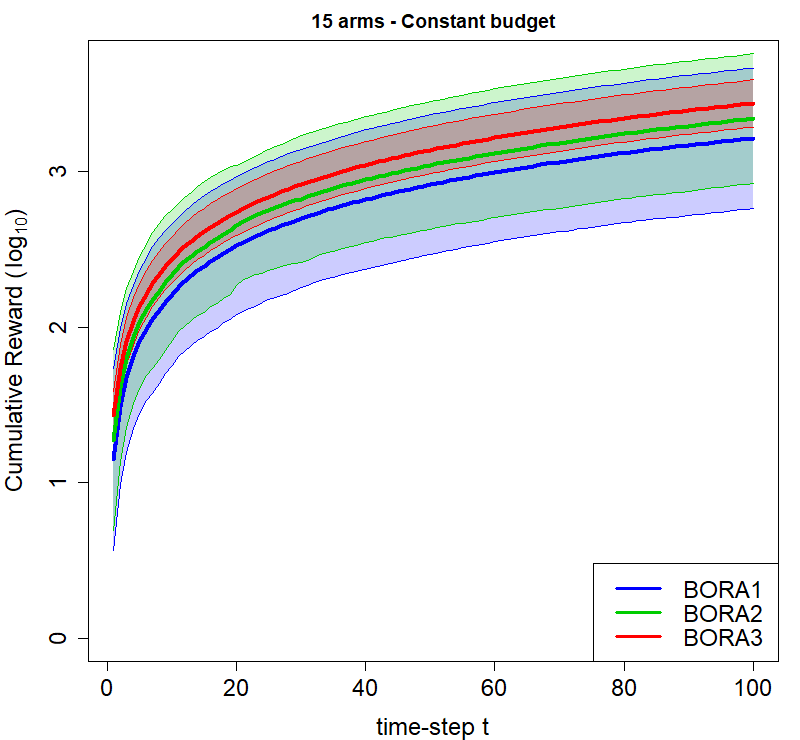}
    \caption{\textbf{Case 2.a} results: cumulative reward (log$_{10}$) of the three BORA algorithms (mean and standard deviation over 5 independent runs).}
    \label{fig:case2a}
\end{figure}

Figure \ref{fig:case2a} shows the cumulative reward of the three BORA algorithms for the \textbf{Case2.a}. On average, BORA$_3$ outperforms the other two algorithms, and BORA$_2$ outperforms BORA$_1$. Moreover, the standard deviation of the cumulative reward, over the 5 independent runs, is significantly lower for BORA$_3$, making it the best approach. More precisely, the cumulative reward at $T=100$, averaged over the 5 independent runs, is $1646.38$, $2198.01$, and $2754.27$, respectively for BORA$_1$, BORA$_2$ and BORA$_3$.

\begin{figure}[h]
    \centering
    \includegraphics[scale=0.35]{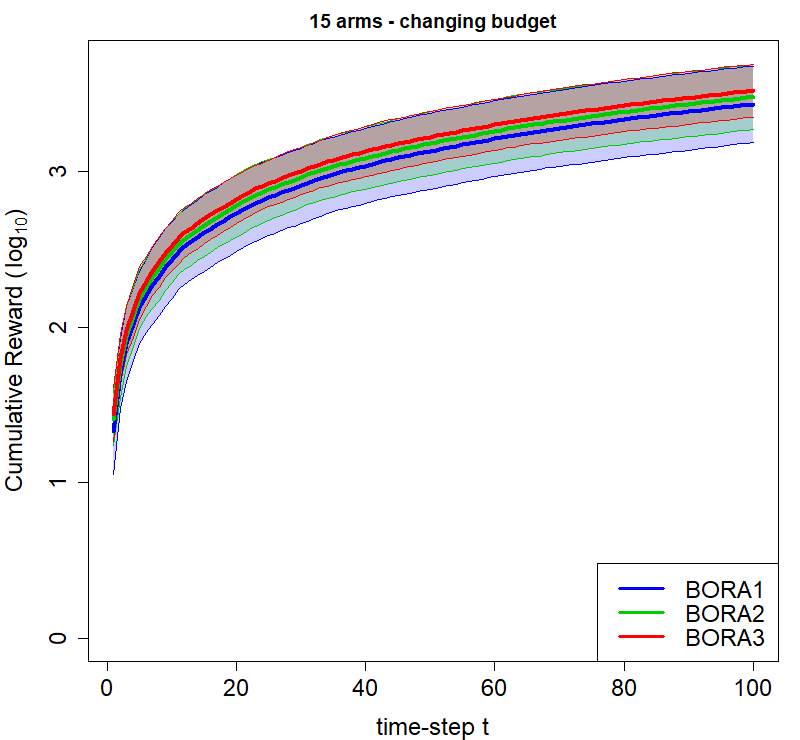}
    \caption{\textbf{Case 2.b} results: cumulative reward (log$_{10}$) of the three BORA algorithms (mean and standard deviation over 5 independent runs).}
    \label{fig:case2b}
\end{figure}

Figure \ref{fig:case2b} shows the cumulative reward of the three BORA algorithms for the \textbf{Case2.b}. Results confirm what observed in the previous case, with BORA$_3$ better, on average, than the other two algorithms, and BORA$_2$ better than BORA$_1$. More precisely, the cumulative reward at $T=100$, averaged over the 5 independent runs, is $2714.27$, $3025.69$, and $3326.237$, respectively for BORA$_1$, BORA$_2$ and BORA$_3$.

\newpage

\section{Conclusions} \label{sec5}
This paper proposes an extension of the optimal resource allocation problem originally proposed in \citep{lattimore2014optimal} to a more general setting. Extensions are specifically related to  (\textit{i}) considering an equality constraint instead of an inequality one (i.e., all the available budget has to be used in making allocation decisions), (\textit{ii}) considering budget changing over time instead of constant, and (\textit{iii}) considering a generic linear immediate reward function instead of one returning the number of successfully completed jobs. All these extensions are motivated by presenting a specific real-life application which cannot be addressed through SBF, that is the optimal budget allocation for multi-channel marketing.

Moreover, empirical results proved that the proposed BORA (Bayesian Optimization for Resource Allocation) approach is more effective and efficient than SBF on the original optimal resource allocation problem. More specifically, BORA shows better generalization capabilities, making it a better learning-and-optimization framework than SBF.

It is important to remark that, while addressing new specific real-life problems (like the one addressed in this paper) has usually required to modify the original SBF algorithm, BORA is more general because it directly learns the immediate reward without considering any other information about its structure. On the contrary, SBF needs to observe the outcome of each decision with respect to every specific job.

Another important conclusion regards the three BORA algorithms proposed. Although all of them are effective and efficient, the one working on probability distributions, through the Wasserstein kernel (i.e., BORA$_3$), was the one showing the best performances, on average, on almost all the caste studies analyzed in the paper.

A possible limitation of the BORA approach is the computational time required to learn the underlying GP model. Indeed, the computational time is significantly higher than SBF. However, this could be a limitation only for settings characterized by a really tight time elapsing between two consecutive time-steps (e.g., lower than one second). For all the other settings, such as the one considered in this paper, the time between $t$ and $t+1$ is largely sufficient to run anyone of the proposed BORA algorithms.

\section*{Code availability}
To guarantee the reputability of the experiments, the code is freely available at the following link:
https://github.com/acandelieri/BORA.git

\newpage

\section*{Appendix}
\begin{figure}[h]
    \centering
    \includegraphics[scale=0.35]{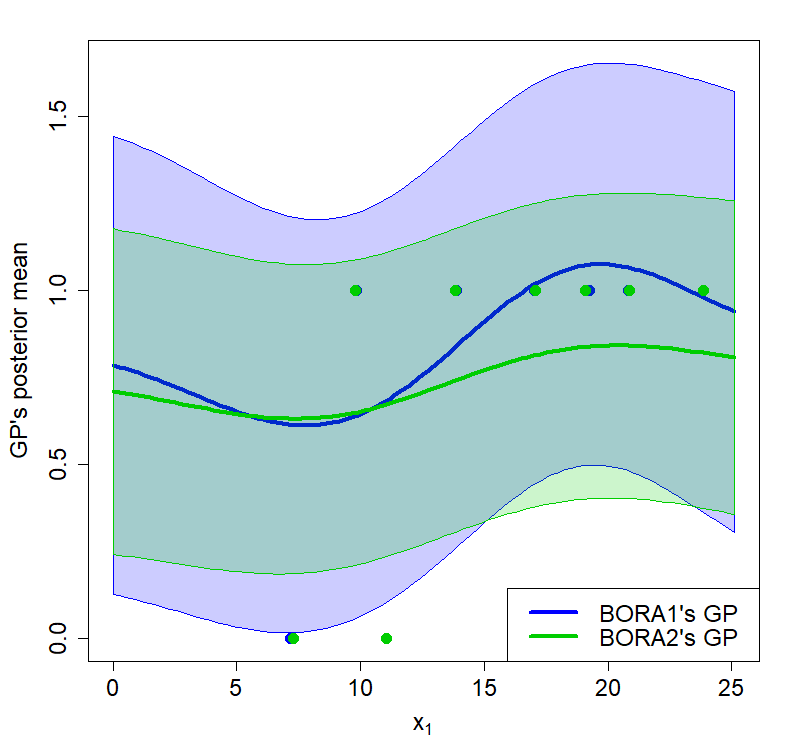}
    \caption{An example of different approximations provided by BORA$_1$'s and BORA$_2$'s GP, given the same set of allocation decisions.}
    \label{fig:extras}
\end{figure}

Figure \ref{fig:extras} shows an example of two different approximations provided by BORA$_1$'s and BORA$_2$'s GP, given the same set of allocation decisions. More precisely, a different estimation of the noise effect leads to two GPs with different smoothness.

\newpage

 \bibliographystyle{elsarticle-num} 
 \bibliography{BORA}





\end{document}